\documentclass[10pt,twocolumn,letterpaper]{article}

\makeatletter
\def\input@path{{ICCV2025-Author-Kit/}} 
\makeatother

\usepackage{placeins}

\usepackage{ICCV2025-Author-Kit/iccv}            

\usepackage[T1]{fontenc}
\usepackage[english]{babel}
\usepackage{csquotes}
\usepackage[nopatch=footnote]{microtype}

\usepackage{amsmath,amsfonts}
\usepackage{mathrsfs}
\usepackage{siunitx}

\usepackage{graphicx}
\graphicspath{{ICCV2025-Author-Kit/}} 
\usepackage{tikz}
\usetikzlibrary{positioning,calc}
\usepackage{pgfplots}
\pgfplotsset{compat=1.18}

\usepackage{booktabs}
\usepackage{threeparttable}
\usepackage{diagbox}
\usepackage{multirow}
\usepackage{tabularx}
\usepackage{array}
\usepackage{makecell}

\usepackage{enumitem}
\usepackage{float}
\usepackage{wrapfig}
\usepackage{needspace}
\usepackage{afterpage}

\usepackage{xcolor}
\definecolor{iccvblue}{rgb}{0.0,0.49,0.69}
\usepackage{listings}

\usepackage[ruled,lined,linesnumbered,commentsnumbered]{algorithm2e}

\usepackage[pagebackref,breaklinks,colorlinks,allcolors=iccvblue]{hyperref}

\usepackage{setspace}
\def\paperID{11358}
\def\confName{ICCV}
\def\confYear{2025}

\title{BEN: Using Confidence-Guided Matting for\\
Dichotomous Image Segmentation}

\author{
Maxwell Meyer \quad Jack Spruyt\\[4pt]
Prama\\[2pt]
{\tt\small \{maxwellmeyer,jackspruyt\}@prama.llc}
}

\microtypesetup{protrusion=true,expansion=true}
\begin{document}
\begin{spacing}{1.23}

\maketitle

\begin{abstract}

Current approaches to dichotomous image segmentation (DIS) treat image matting and object segmentation as fundamentally different tasks. As improvements in image segmentation become increasingly challenging to achieve, combining image matting and grayscale segmentation techniques offers promising new directions for architectural innovation. Inspired by the possibility of aligning these two model tasks, we propose a new architectural approach for DIS called Confidence-Guided Matting (CGM). We created the first CGM model called Background Erase Network (BEN). BEN consists of two components: BEN Base for initial segmentation and BEN Refiner for confidence-based refinement. Our approach achieves substantial improvements over current state-of-the-art methods on the DIS5K validation dataset, demonstrating that matting-based refinement can significantly enhance segmentation quality. This work introduces a new paradigm for integrating matting and segmentation techniques, improving fine-grained object boundary prediction in computer vision.

\end{abstract}

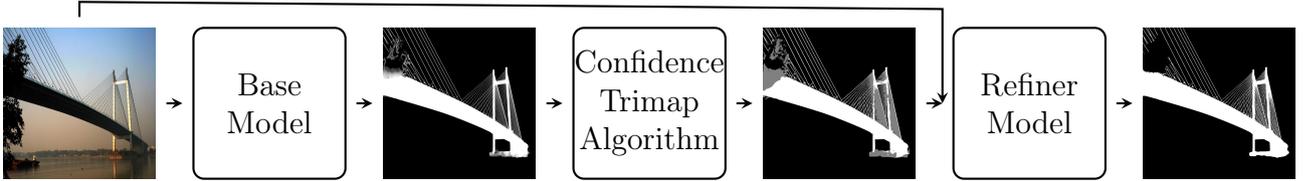
\begin{figure*}[t]
  \centering
  \begin{tikzpicture}[
      >=stealth,
      node distance=2.5cm,
      baseline=(current bounding box.center)
  ]
  \node (input) {\includegraphics[width=2cm,height=2cm]{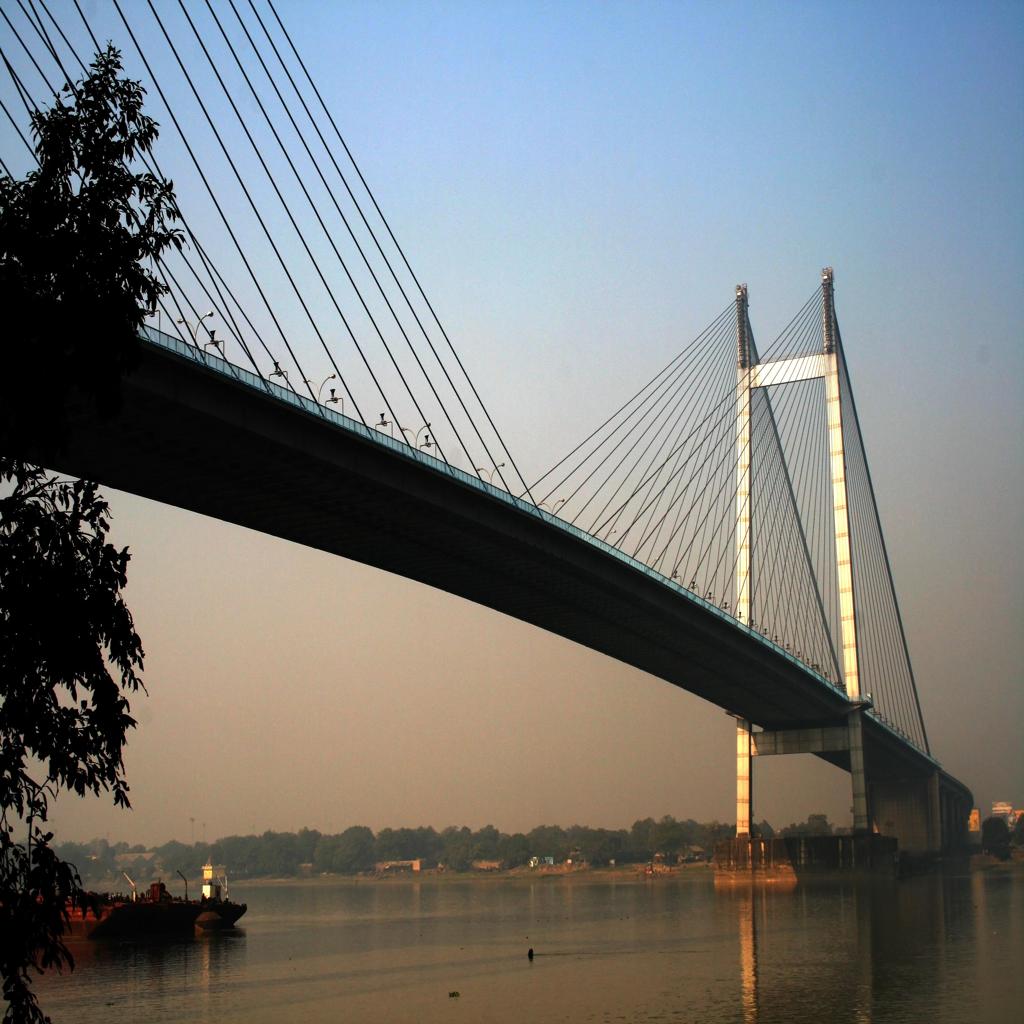}};
  \node (baseBox) [right of=input] {%
    \begin{tikzpicture}[baseline=(boxText.center)]
      \path[use as bounding box] (0,0) rectangle (2,2);
      \draw[rounded corners=4pt, thick] (0,0) rectangle (2,2);
      \node (boxText) at (1,1) {\parbox{1.8cm}{\centering Base Model}};
    \end{tikzpicture}
  };
  \node (baseMask) [right of=baseBox]
    {\includegraphics[width=2cm,height=2cm]{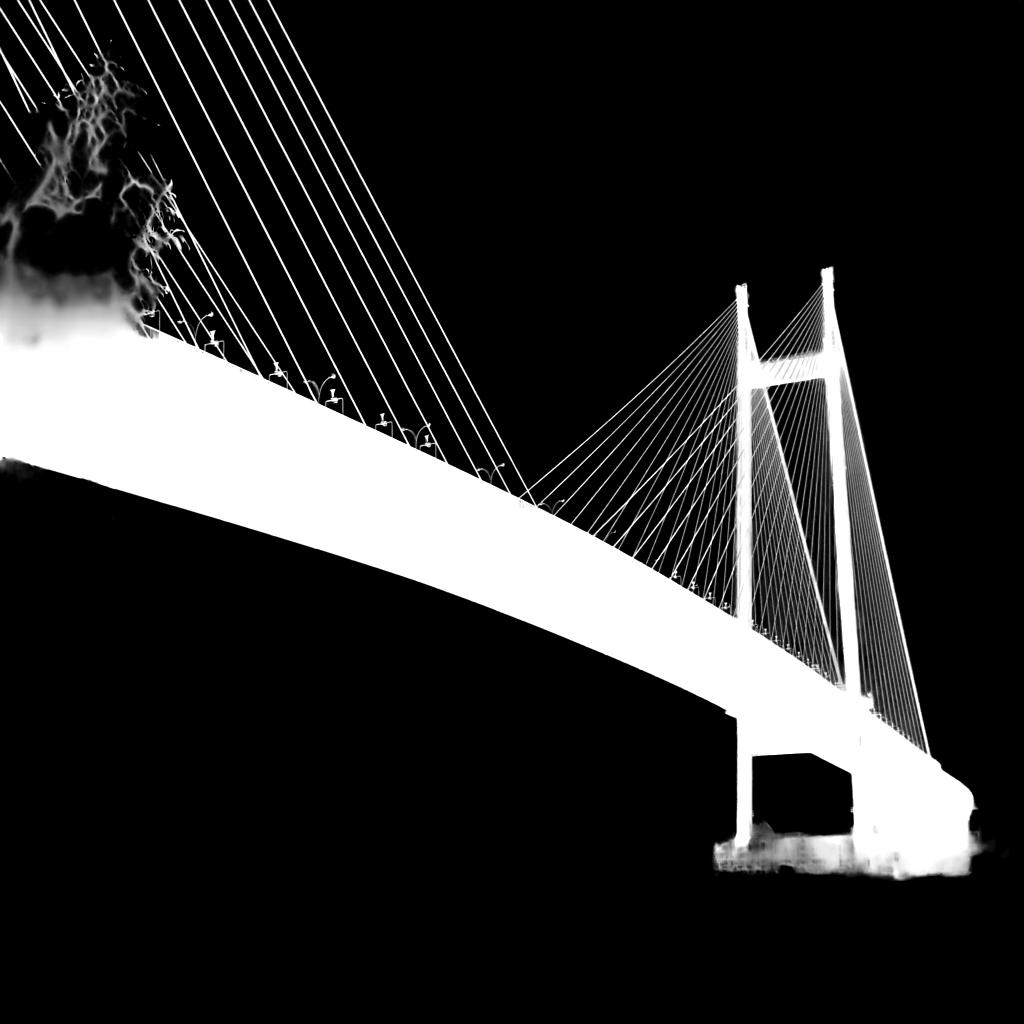}};
  \node (trimapBox) [right of=baseMask] {%
    \begin{tikzpicture}[baseline=(boxText.center)]
      \path[use as bounding box] (0,0) rectangle (2,2);
      \draw[rounded corners=4pt, thick] (0,0) rectangle (2,2);
      \node (boxText) at (1,1)
        {\parbox{2cm}{\centering Confidence\\Trimap\\Algorithm}};
    \end{tikzpicture}
  };
  \node (trimap) [right of=trimapBox]
    {\includegraphics[width=2cm,height=2cm]{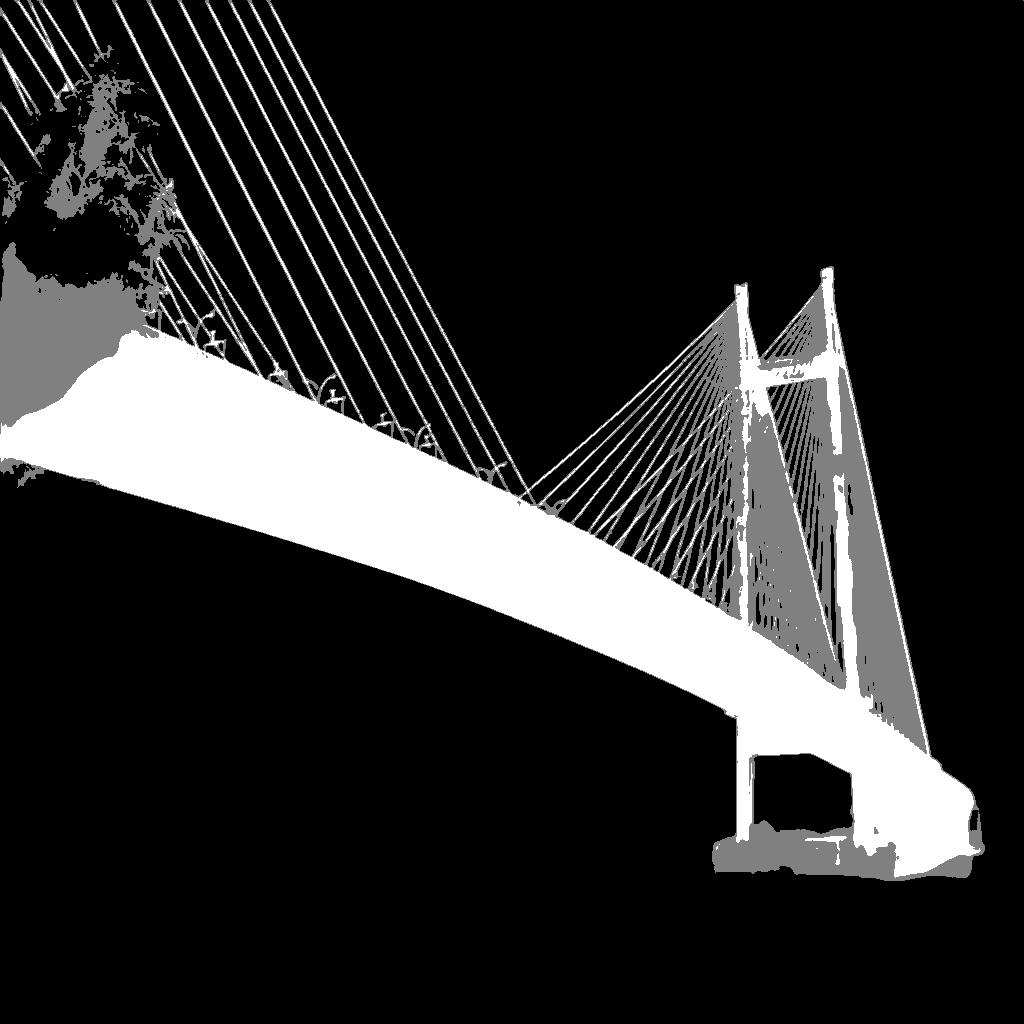}};
  \node (refinerBox) [right of=trimap] {%
    \begin{tikzpicture}[baseline=(boxText.center)]
      \path[use as bounding box] (0,0) rectangle (2,2);
      \draw[rounded corners=4pt, thick] (0,0) rectangle (2,2);
      \node (boxText) at (1,1) {\parbox{1.8cm}{\centering Refiner Model}};
    \end{tikzpicture}
  };
  \node (finalMask) [right of=refinerBox]
    {\includegraphics[width=2cm,height=2cm]{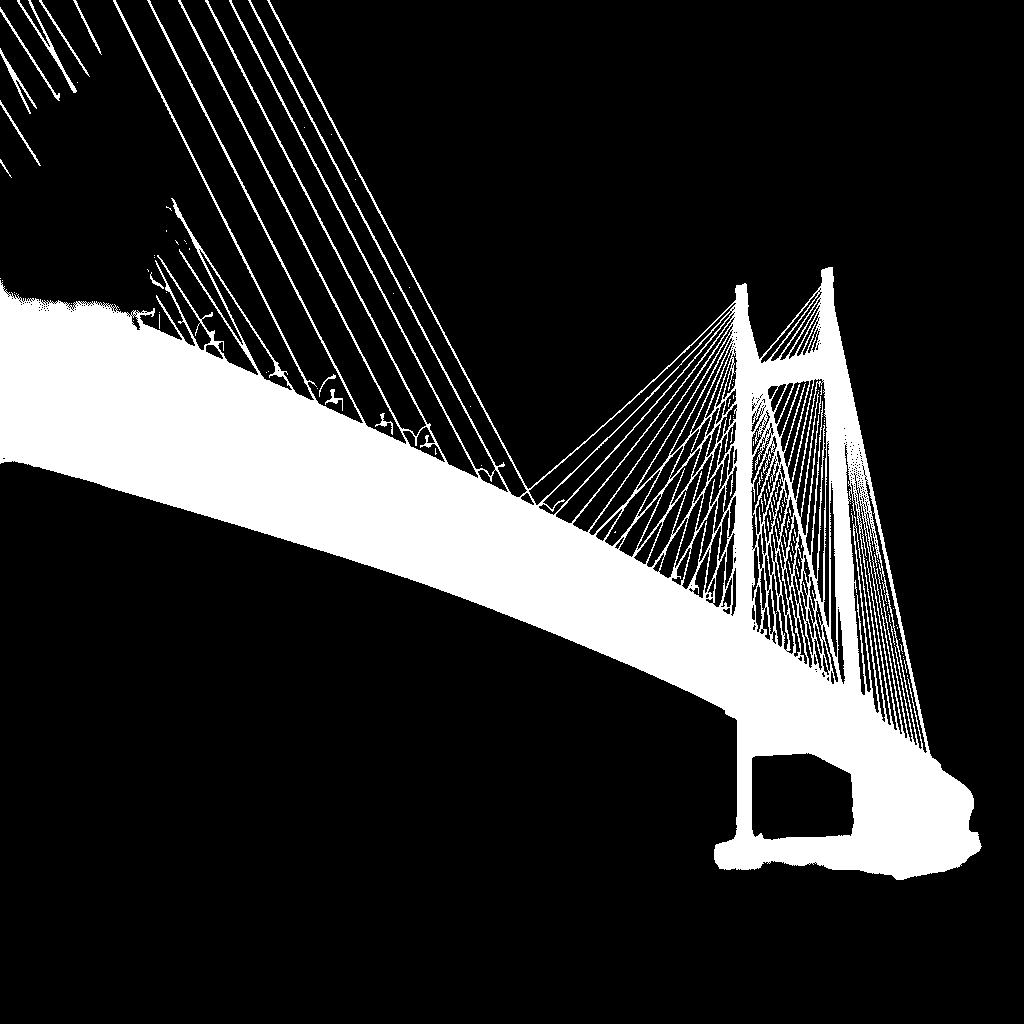}};
  \draw[->,thick] (input.east)       -- (baseBox.west);
  \draw[->,thick] (baseBox.east)     -- (baseMask.west);
  \draw[->,thick] (baseMask.east)    -- (trimapBox.west);
  \draw[->,thick] (trimapBox.east)   -- (trimap.west);
  \draw[->,thick] (trimap.east)      -- (refinerBox.west);
  \draw[->,thick] (refinerBox.east)  -- (finalMask.west);
  \draw[->,thick]
    (input.north)             
    -- ++(0,0.2cm)
    -- ($(refinerBox.west)+(0,1.25cm)$) 
    -- (refinerBox.west);     
  \end{tikzpicture}
  \caption{\rm Shows the process flow from input image through Base Model, Confidence Trimap Algorithm, and Refiner Model stages to produce the final segmentation mask.}
  \label{fig:cgm_pipeline}
\end{figure*}

\section{Introduction}
\label{sec:intro}

Image segmentation is used in a wide range of tasks, including autonomous vehicle navigation~\cite{choi2020cars,che2023twinlitenet} and medical imaging~\cite{Wu_2024_CVPR,Zhang_2024_CVPR,10030099}. Dichotomous image segmentation (DIS) is the segmentation of foreground objects from real images. The DIS5K~\cite{qin2022highly} dataset contains 5,470 high-resolution images and their corresponding pixel-accurate masks. This dataset is comprised of a wide range of diverse categories, from musical instruments to automobiles. The DIS5K dataset has become a vital benchmark for understanding segmentation performance in real-world scenarios. 

Recent efforts have led to significant accuracy improvements on the DIS5K dataset through novel architectures. The Multi-view Aggregation Network (MVANet)~\cite{yu2024multi} achieved state-of-the-art results using multi-view learning and a Swin backbone~\cite{liu2021swin}. However, transformer-based approaches face limitations due to the computational cost of attention in Vision Transformer (ViT)~\cite{dosovitskiy2020image} and Swin backbone architectures. Additionally, these limitations make batch edge prediction impractical. When encountering data distributions significantly different from their training set, these ViT and Swin backbone models often display reduced confidence in foreground-background classification, leading to unwanted matting effects. The DiffDIS~\cite{yu2024high} demonstrated that incorporating edge prediction alongside foreground segmentation improves the accuracy of the final output. DiffDIS employs a Variational Auto Encoder (VAE)~\cite{kingma2013auto} in order to enable the diffusion model to operate on high-resolution images in a compressed latent space before they are reconstructed to their input resolution. This approach supports scalable batch sizes, facilitating precise edge and foreground prediction.

We propose Confidence-Guided Matting (CGM), an architectural approach that bridges the gap between segmentation and matting, enabling improved segmentation performance. Current matting approaches generate trimaps manually. These trimaps are used to predict fine-grained details, focusing on regions where the foreground meets the background. Trimaps consist of three regions: black (background), white (foreground), and gray (unknown). The gray area of the trimap is updated with the foreground or background prediction by a matting model.

Our method innovates on existing matting techniques by dynamically generating trimaps based on the confidence levels from the base model's predictions. These confidence trimaps allow a refiner network to focus on uncertain areas, where the foreground transitions into the background or regions with complex, fine-grained details. Our Background Erase Network (BEN) incorporates this approach by employing a base network for initial segmentation and a refiner network for confidence-guided refinement. Together, these components work in concert, as depicted in Figure~\ref{fig:cgm_pipeline}, to achieve new state-of-the-art accuracy on the DIS5K validation dataset (DIS-VD).

\section{Related Works}
\label{sec:relatedworks}

\subsection{Advances in Dichotomous Image Segmentation}
Dichotomous image segmentation (DIS) can be described as highly accurate segmentation of complex foreground objects from their backgrounds in real-world images. The DIS5K dataset~\cite{qin2022highly} is a high-resolution non-conflicting dataset built for DIS. This dataset is currently the most critical dataset when seeking to describe the performance of a pixel-wise accurate segmentation model.

The Multi-view Aggregation Network (MVANet)~\cite{yu2024multi} has shown significant promise in DIS by utilizing a multi-view learning approach. The MVANet separates the images into global and local patches, which are concatenated batch-wise and then inputted to a Swin Transformer~\cite{liu2021swin} backbone. The Swin Transformer generates hierarchical representations of each of the patches, which then share information with one another using Multi-view Complementary Refinement and Multi-view Complementary Localization blocks, which are up-sampled in the process. This process ultimately aggregates the information into one final output. This architectural approach shows state-of-the-art performance at a modest 94 million parameters. We expand the core ideas of the MVANet architecture to create our BEN Base model.

Diffusion DIS (DiffDIS)~\cite{yu2024high} attempts to shift to a diffusion-based approach, as diffusion has played a crucial role in computer vision ~\cite{rombach2022high}. The DiffDIS allows for a multi-batch estimation of the segmentation and edge prediction. Edge prediction alongside segmentation significantly increases the detail of the foreground's edges. Along with other innovations, this DiffDIS model surpasses the MVANet model on the DIS5K validation dataset. The key contribution from the DiffDIS to our approach is the insight that fine-grained edge accuracy will increase when predicting the edges of any given foreground scene. This idea is one of our key motivations for CGM approach.

\subsection{Deep Learning Image Matting}
Image matting leverages alpha mattes to discern the foreground from the background. The ambiguous areas are often complicated scenes such as hair or smoke. There has been a shift from convolution neural network approaches to transformer-based architectures. Vision Transformer-based matting models have shown state-of-the-art results by incorporating attention mechanisms~\cite{yao2024vitmatte}. The models show great efficacy in using trimaps to guide foreground refinement.
These techniques pave the way for confidence-based trimap generation, hence our work unifying them with CGM on the DIS5K dataset.

\section{Method}

 \subsection{BEN Base} The BEN Base architecture resembles that of the MVANet but with notable changes. We modified the activation function and normalization when redesigning the MVANet for our base model. MVANet was originally trained with a batch size of one; therefore, we use instance normalization instead of batch normalization in BEN Base, as batch normalization is known to be unstable for small batch sizes~\cite{ulyanov2016instance}. We adopt Gaussian Error Linear Unit ~\cite{hendrycks2016gaussian} as the activation function rather than the Rectified Linear Unit and Parametric Rectified Linear Unit used in the original MVANet. 

\subsection{Loss Function}We use three metrics for BEN Base's loss function: weighted BCE (WBCE), weighted IoU (WIOU), and weighted structural similarity (SSIM). We define a \emph{Structure Loss} to measure the difference between a predicted segmentation (logits) and its ground-truth mask. Our loss closely resembles that of the BiRefNet~\cite{zheng2024bilateral}. This loss leverages all three of the previous equations. Formally written as: \begin{equation} \begin{split} \text{StructureLoss}(p, y) &= w_{\text{BCE}} \cdot \text{WBCE}(p, y) \\ &\quad + w_{\text{IOU}} \cdot \text{WIOU}(\sigma(p), y) \\ &\quad + w_{\text{SSIM}} \cdot \text{SSIM}(\sigma(p), y), \end{split} \end{equation} where \(y\) denotes the ground truth mask, \(p\) denotes predicted logits, and \(\sigma(p)\) denotes their sigmoid. The \(w_{\text{BCE}}, w_{\text{IOU}}, w_{\text{SSIM}}\) are set to four, one, and two, respectively, for all of the experiments. Each loss component (local, global, and token) sums their StructureLoss terms across various scales to capture details at multiple resolutions.
\begin{align}
\mathcal{L}_{\text{local}} &= \sum_{i=1}^{6} \text{StructureLoss}(s_i, y_i), \\
\mathcal{L}_{\text{global}} &= \sum_{i=1}^{5} \text{StructureLoss}(g_i, y_i), \\
\mathcal{L}_{\text{token}} &= \sum_{i=1}^{4} \text{StructureLoss}(t_i, y_i)
\end{align}
where \(s_i\) are side outputs and includes the final output, \(g_i\) are global outputs. where \(t_i\) are token outputs at different scales. Building on these scale-specific terms, we define a \emph{combined} loss that balances local, global, and token-level predictions. This \emph{normal combined loss} is then defined by: \begin{equation} \mathcal{L}_{\text{combined}} = 0.3 \,\mathcal{L}_{\text{global}} \;+\; 0.3 \,\mathcal{L}_{\text{token}} \;+\; \mathcal{L}_{\text{local}} \end{equation}

\begin{algorithm}[tbp]   
\DontPrintSemicolon
\SetAlgoLined
\SetKwInOut{Input}{Input}
\SetKwInOut{Output}{Output}
\caption{Trimap Generation}\label{algorithm:trimap}
\Input{Predicted mask $P$, $H=W=1024$}
\Output{Trimap $T\in\{0,128,255\}^{H\times W}$}
$t_h \gets 0.95,\; t_l \gets 0.05,\; M \gets \sigma(P)$\;

\ForEach{$(i,j)\in H\times W$}{
  \If{$M_{i,j}\ge t_h$}{$T_{i,j}\gets 255$}
  \ElseIf{$M_{i,j}\le t_l$}{$T_{i,j}\gets 0$}
  \Else{$T_{i,j}\gets 128$}
}
\end{algorithm}
\vspace*{6mm}  

\subsection{Confidence-Guided Matting} To address the limitations of matting's current application in DIS techniques, we propose Confidence-Guided Matting (CGM). CGM allows a base prediction to be updated depending on the model's confidence that a given pixel is a foreground or background pixel. Unlike traditional trimaps, which require manual labeling or predefined thresholds, our trimaps are dynamically generated based on the base model’s confidence in each pixel’s classification. The refiner network subsequently learns how to optimize the uncertain regions, which leads to improved segmentation accuracy. To integrate the base and refiner model, we generate confidence trimaps from the base model's sigmoid prediction. These confidence trimaps are then passed to the refiner for the final prediction. The trimap generation process is formally presented in Algorithm~\ref{algorithm:trimap}.

\begin{figure*}[tbp]
  \scalebox{1.0}{
    \begin{minipage}{\textwidth}
      \centering
      \includegraphics[width=\linewidth]{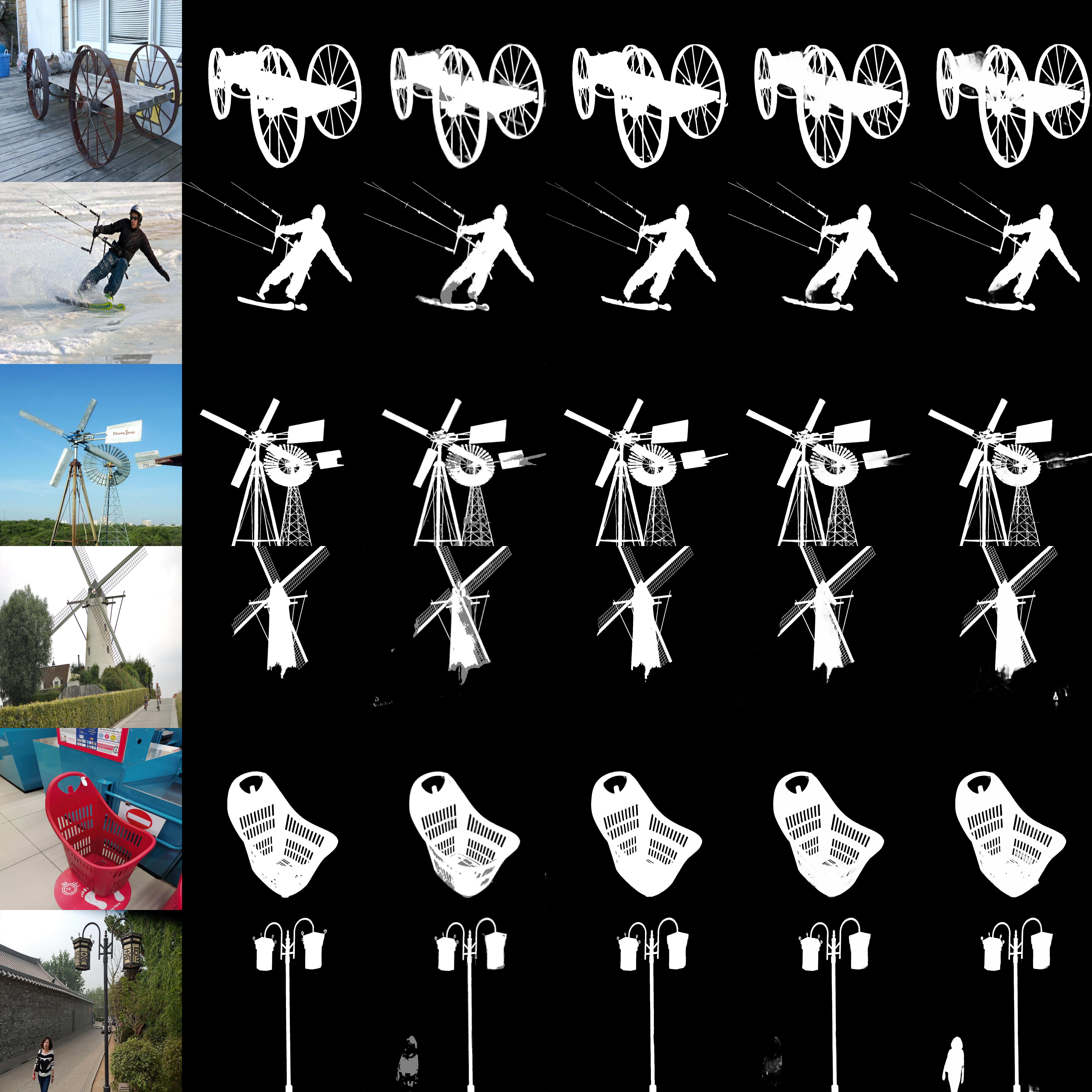}

      \renewcommand{\arraystretch}{1.0}
      \begin{tabularx}{\linewidth}{*{6}{>{\centering\arraybackslash}X}}
        \makecell{Input} & \makecell{GT} &
        \makecell{Confidence \\ Trimap} &
        \makecell{BEN Base+ \\ Refiner} &
        \makecell{BEN Base} & \makecell{MVANet}
      \end{tabularx}
    \end{minipage}%
  }
  \caption{Qualitative results from DIS5K validation dataset}
  \label{fig:2}
\end{figure*}


\begin{table*}[tbp]
\centering
\resizebox{\textwidth}{!}{%
\begin{tabular}{l|ccccc|cc}
\hline
\multicolumn{8}{c}{DIS-VD} \\
\hline
Method & InSPyReNet & BiRefNet & MVANet & GenPercept & DiffDIS & BEN\_Base & BEN\_Base+Refiner \\
\hline
$F_{\beta}^{max}\uparrow$ & 0.889 & 0.897 & 0.913 & 0.844 & 0.918 & \textbf{0.923} & 0.919 \\
$F_{\beta}^{\omega}\uparrow$ & 0.834 & 0.863 & 0.856 & 0.824 & 0.888 & 0.871 & \textbf{0.896} \\
$E_{\phi}^m\uparrow$ & 0.914 & 0.937 & 0.938 & 0.924 & 0.948 & 0.935 & \textbf{0.959} \\
$S_m\uparrow$ & 0.900 & 0.905 & 0.905 & 0.863 & 0.904 & 0.916 & \textbf{0.917} \\
$\mathcal{M}\downarrow$ & 0.042 & 0.036 & 0.036 & 0.044 & 0.029 & 0.031 & \textbf{0.027} \\
\hline
\end{tabular}}
 \caption{Quantitative comparison of DIS5K validation dataset with common representative methods to compare our models to existing DIS models. $\uparrow$ represents "higher is better," and $\downarrow$ represents "lower is better". The best score is indicated in \textbf{bold}. External scores are from~\cite{yu2024high}.} 
\label{tab:quantitative_comparison}
\end{table*}

\vspace*{3mm}

\section{Experiments}
\vspace*{2mm}

\subsection{Data and Evaluation Metrics}
To leverage both commercial interest and the open-source community, we train on the DIS5K dataset and preserve the DIS5K validation set for fair comparison with other publicly available models. We evaluate our results using widely adopted metrics as follows: Max F-measure ($F_{\beta}^{max}$)~\cite{6247743}, Weighted F-measure ($F_{\beta}^{\omega}$)~\cite{Margolin_2014_CVPR}, E-measure ($E_{\phi}^m$)~\cite{fan2018enhanced}, Structural Similarity Measure ($S_{m}$)~\cite{fan2017structure}, Mean Absolute Error (MAE, $\mathcal{M}$)~\cite{6247743}, Dice Coefficient (Dice)~\cite{shamir2019continuous}, Intersection over Union (IoU)~\cite{rezatofighi2019generalized}, Balanced Error Rate (BER)~\cite{liu2022ber}, and Accuracy (Acc).

\begin{table}[htbp]
\centering
\scriptsize
\resizebox{\linewidth}{!}{%
\begin{tabular}{l|ccc}
\hline
\multicolumn{4}{c}{DIS-VD} \\
\hline
Method & MVANet & BEN\_Base & BEN\_Base+Refiner \\
\hline
MAE$\downarrow$ & 0.036 & 0.031 & \textbf{0.027} \\
Dice$\uparrow$   & 0.869 & 0.881 & \textbf{0.899} \\
IoU$\uparrow$    & 0.807 & 0.837 & \textbf{0.851} \\
BER$\downarrow$  & 0.065 & 0.052 & \textbf{0.049} \\
Acc$\uparrow$    & 0.966 & 0.972 & \textbf{0.974} \\
\hline
\end{tabular}
}
\caption{Comprehensive comparison of our BEN models with MVANet.}
\label{tab:comprehensive_comparison}
\end{table}

 \subsection{Quantitative Results} To evaluate our model, we examined the data from 5 different models, InSPyReNet~\cite{kim2022revisiting}, BiRefNet~\cite{zheng2024bilateral}, MVANet~\cite{yu2024multi}, GenPercept~\cite{xu2024diffusion} and DiffDIS~\cite{yu2024high} Table~\ref{tab:quantitative_comparison}. The BEN family occupies the top slot for each of the DIS 5k measurement methods. Based on the ($F_{\beta}^{max}$) metric, BEN Base alone slightly outperforms BEN Base+Refiner. To perform a deeper examination of our performance, we expanded our evaluation with additional metrics and used MVANet as a control depicted in Table~\ref{tab:comprehensive_comparison}. BEN Base + Refiner significantly outperforms MVANet and BEN Base as well. 


\subsection{Qualitative Results}
To intuitively understand how the refiner is updating the base predictions we selected outputs from the confidence trimap algorithm, BEN Base, and BEN Base + Refiner that are in the training dataset. As shown in Figure~\ref{fig:2}, the Refiner is correctly able to remove and refine edges of the base models prediction. This combination significantly outperforms the MVANet as it is unable to correctly define foreground objects and attend to fine details.

\subsection{Ablation Test}
\begin{table}[h]
\centering
\resizebox{0.95\columnwidth}{!}{%
\begin{tabular}{cc|ccccc}
\hline
\multicolumn{7}{c}{DIS-VD} \\
\hline
Low & High & MAE & Dice & IoU & BER & Acc \\
\hline
0.45 & 0.55 & 0.029 & 0.895 & 0.842 & 0.049 & 0.972 \\
0.35 & 0.65 & 0.028 & 0.896 & 0.845 & 0.050 & 0.973 \\
0.25 & 0.75 & 0.028 & 0.897 & 0.847 & 0.050 & 0.973 \\
0.15 & 0.85 & \textbf{0.027} & 0.898 & 0.849 & 0.050 & \textbf{0.974} \\
0.05 & 0.95 & \textbf{0.027} & \textbf{0.899} & \textbf{0.851} & \textbf{0.049} & \textbf{0.974} \\
0.01 & 0.99 & \textbf{0.027} & \textbf{0.899} & \textbf{0.851} & 0.050 & \textbf{0.974} \\
0.005 & 0.995 & \textbf{0.027} & 0.897 & 0.848 & 0.051 & \textbf{0.974} \\
\hline
\end{tabular}}
\caption{Comprehensive comparison of various refinement settings for BEN\_Base+Refiner on the DIS-VD. $\uparrow$ indicates "higher is better" and $\downarrow$ indicates "lower is better". The best scores are indicated in \textbf{bold}.} 
\label{tab:new_confidence_table}
\end{table}

We examine the impact of modifying the low and high confidence thresholds in the trimap generation. In Table~\ref{tab:new_confidence_table}, the comparison of the thresholds shows the efficacy of the proposed 0.05 and 0.95 thresholds. These thresholds show the best trade-off between the refiner's control of the output versus the base prediction's control.

\section{Conclusion} In this paper, we examine the union of image foreground segmentation and image matting with the creation of BEN. We leverage CGM to dynamically generate confidence trimaps that allow the Refiner sufficient control over uncertain areas from BEN Base's initial prediction. Automating the trimap generation enables the Refiner to learn uncertain regions. This powerful combination shows significant improvements over existing segmentation methods on the DIS5K validation dataset. This introduces a novel approach to segmentation, unifying matting with high-precision segmentation. We plan to extend this work to other image segmentation tasks in future work. Our work supports further exploration of confidence-based matting techniques in precise image segmentation. 

\end{spacing}

{\small
\bibliographystyle{ICCV2025-Author-Kit/ieeenat_fullname}
\bibliography{ICCV2025-Author-Kit/main}
}

\end{document}